\documentclass[10pt,twocolumn,letterpaper]{article}

\usepackage{iccv}
\usepackage{times}
\usepackage{graphicx}
\usepackage{amsmath}
\usepackage{amssymb}

\usepackage[pagebackref=true,breaklinks=true,letterpaper=true,colorlinks,bookmarks=false]{hyperref}

\iccvfinalcopy 

\def\iccvPaperID{30} 

\ificcvfinal\fi

\begin{document}

\title{Better Exploiting OS-CNNs for Better Event Recognition in Images}

\author{Limin Wang \quad \quad Zhe Wang  \quad \quad Sheng Guo \quad \quad Yu Qiao \\
Shenzhen Key Lab of CVPR,  Shenzhen Institutes of Advanced Technology, CAS, China \\
{\tt\small \{07wanglimin, buptwangzhe2012, guosheng1001\}@gmail.com, yu.qiao@siat.ac.cn}}

\maketitle

\begin{abstract}
Event recognition from still images is one of the most important problems for image understanding. However, compared with object recognition and scene recognition, event recognition has received much less research attention in computer vision community. This paper addresses the problem of cultural event recognition in still images and focuses on applying deep learning methods on this problem. In particular, we utilize the successful architecture of \emph{Object-Scene Convolutional Neural Networks} (OS-CNNs) to perform event recognition. OS-CNNs are composed of object nets and scene nets, which transfer the learned representations from the pre-trained models on large-scale object and scene recognition datasets, respectively. We propose four types of scenarios to explore OS-CNNs for event recognition by treating them as either ``end-to-end event predictors'' or ``generic feature extractors''. Our experimental results demonstrate that the global and local representations of OS-CNNs are complementary to each other. Finally, based on our investigation of OS-CNNs, we come up with a solution for the cultural event recognition track at the ICCV ChaLearn Looking at People (LAP) challenge 2015. Our team secures the third place at this challenge and our result is very close to the best performance.
\end{abstract}

\section{Introduction}
\label{sec:intro}

\begin{figure}
 \includegraphics[width=\linewidth]{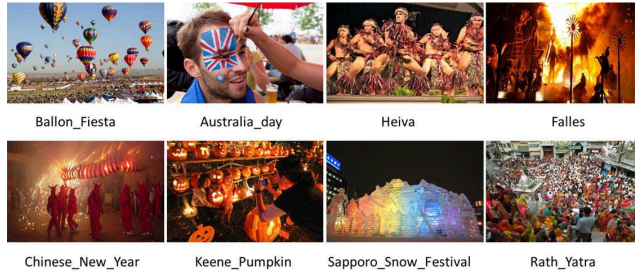}
 \caption{Examples of cultural event images from the ICCV ChaLearn Looking at People (LAP) dataset. From these examples, we can see that the characterization of event is complicated and it is related to many visual cues, such as objects, secne category, and human garments.}
 \label{fig:example}
\end{figure}

Image understanding \cite{KrizhevskySH12,SimonyanZ14a,SzegedyLJSRAEVR14,ZeilerF14} is becoming one of the most important problems in computer vision and many research efforts have been devoted to this topic. While object recognition \cite{DengDSLL009} and scene recognition \cite{ZhouLXTO14} have been extensively studied in the task of image classification, event recognition \cite{LiF07,WangWDQ15,XiongZLT15} in still images received much less research attention, which also plays an important role in semantic image interpretation. As shown in Figure \ref{fig:example}, the characterization of event is extremely complicated as the event concept is highly related to many other high-level visual cues, such as objects, scene categories, human garments, human poses, and other context. Therefore, event recognition in still images poses more challenges for the current state-of-the-art image classification methods, and needs to be further investigated in the computer vision research.

Convolutional neural networks (CNNs) \cite{lecun-98} have recently enjoyed great successes in large-scale image classification, in particular for object recognition \cite{HeZR015,SimonyanZ14a,SzegedyLJSRAEVR14} and scene recognition \cite{WangGH015,ZhouLXTO14}. For event recognition, much fewer deep learning methods have been designed for this problem. Our previous work \cite{WangWDQ15} proposed a new deep architecture,  called \emph{Object-Scene Convolutional Neural Network} (OS-CNN), for cultural event recognition. OS-CNNs are designed to extract useful information for event understanding from the perspectives of containing objects and scene categories, respectively. OS-CNNs are composed of two-stream CNNs, namely object nets and scene nets. Object nets are pre-trained on the large-scale object recognition datasets (e.g. ImageNet \cite{DengDSLL009}), and scene nets are based on models learned from the large-scale scene recognition datasets (e.g. Places205 \cite{ZhouLXTO14}). Decomposing into object nets and scene nets enables us to use the external large-scale annotated images to initialize OS-CNNs, which may be further fine tuned elaborately on the event recognition dataset. Finally, event recognition is performed based on the late fusion of softmax outputs of object nets and scene nets.

\begin{figure*}
  \includegraphics[width=\textwidth]{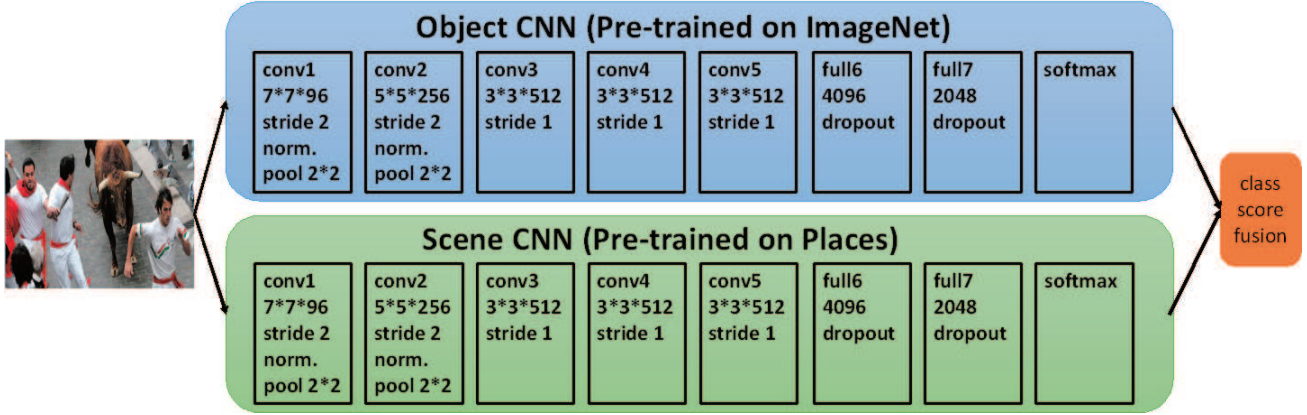}
  \caption{The architecture of Object-Scene Convolutional Neural Network (OS-CNN) for event recognition from \cite{WangWDQ15}.  OS-CNN is composed to two-stream networks: object nets and scene nets, which are separately pre-trained on the ImageNet and Places205 dataset.}
  \label{fig:os-cnn}
\end{figure*}

Following the research line of OS-CNNs, in this paper, we try to further explore different aspects of OS-CNNs and better exploit OS-CNNs for better event recognition. Specifically, we design four types of investigation scenarios to study the performance of OS-CNNs. In the first scenario, we directly use the softmax outputs of CNNs as recognition results. In the next three scenarios, we treat CNNs as feature extractors, and use them to extract both \emph{global} and \emph{local} features of an image region. Global features are more compact and aim to capture the holistic structure, while local features focus on describing the image details and local patterns. Our experimental results indicate these two kinds of features are complementary to each other and robust for event recognition. Based on our empirical explorations with OS-CNNs, we come up with our solution for the cultural event recognition track at the ICCV ChaLearn Looking at People (LAP) challenge \cite{LAP15} and we secure the third place.

The rest of this paper is organized as follows. In Section \ref{sec:os-cnn}, we will give a brief introduction to OS-CNNs, including network architectures and implementation details. After that, we will introduce our extensive explorations with OS-CNNs for event recognition in Section \ref{sec:feature}. We then report our experimental results in Section \ref{sec:exp}. Finally, we conclude our method and present the future work in Section \ref{sec:con}.

\section{OS-CNNs Revisited}
\label{sec:os-cnn}

In this section, we will first briefly introduce the architecture of \emph{Object-Scene Convolutional Neural Networks} (OS-CNNs), which was proposed in our previous work \cite{WangWDQ15}. Then, we will present the implementation details of OS-CNNs, including network structures, data augmentations, and learning policy.

\subsection{OS-CNNs}
Event is a relatively complicated concept in computer vision research and highly related with other two problems: object recognition and scene recognition. The basic idea behind OS-CNN is to utilize two separate components to perform event recognition from the perspectives of occurring objects and scene context. Specifically, OS-CNNs are composed of object nets and scene nets, as shown in Figure \ref{fig:os-cnn} .

\textbf{Object nets.} Object net is designed to capture useful information of objects to help event recognition. Intuitively the occurring objects are able to provide useful cues for event understanding. For instance, in the cultural event of Australia Day as shown in Figure \ref{fig:example},  Australian flag will be a representative object. As the main goal of object net is to deal with object cues, we build it based on recent advances on large-scale object recognition, and pre-train the network on the public ImageNet models. Then, we further fine tune the model parameters on the training dataset of cultural event recognition by setting the output number as $100$ (cultural event recognition dataset containing 100 classes).

\textbf{Scene nets.} Scene net is expected to extract scene information of image to assist event understanding. In general, the scene context will be helpful for recognizing the event category in the image. For example, in the cultural event of Sapporo Snow Festival as shown in Figure \ref{fig:example}, outdoor will be usually the scene category. Specifically, we pre-train the scene nets by using the models learned on the dataset Places205, which contains 205 scene classes and 2.5 millions images. Similar to object nets, we then fine tune the network weights of scene nets on the event recognition dataset, where we set network output number as $100$.

Based on the above analysis, recognizing cultural event will benefit from the transferred representations learned for object recognition and scene recognition. Thus, we will fuse the network outputs of both object nets and scene nets as the prediction of OS-CNNs.

\begin{figure*}[t]
  \includegraphics[width=\textwidth]{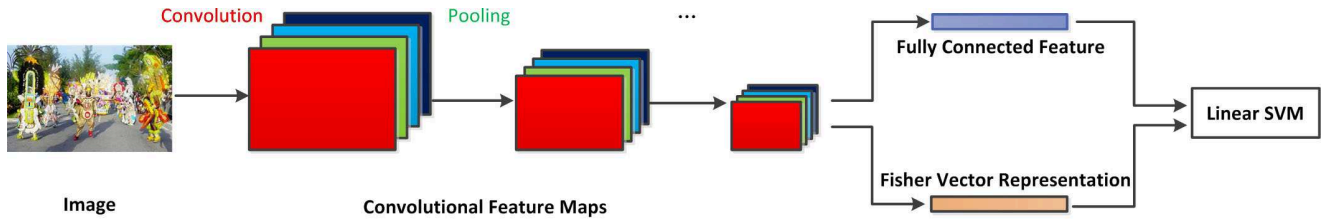}
  \caption{Our better explorations with OS-CNNs for event recognition. We utilize OS-CNNs to extract both global representations (activations of fully connected layers) and local representations (activations of convolutional layers), which can be combined for event recognition in still images.}
  \label{fig:pipeline}
\end{figure*}
\subsection{Implementation details}
In this subsection, we will describe the implementation details of training OS-CNNs, including network structures,  data augmentations, and learning policy.

\textbf{Network structures.}  Network structures are of great importance for improving the performance of CNNs. In the past several years, many successful network architectures have been proposed for object recognition, such as AlexNet \cite{KrizhevskySH12}, ClarifaiNet \cite{ZeilerF14}, OverFeat \cite{SermanetEZMFL13}, GoogLeNet \cite{SzegedyLJSRAEVR14}, VGGNet \cite{SimonyanZ14a}, MSRANet \cite{HeZR015}, and Inception2 \cite{IoffeS15}. Some good practices can be drawn from the evolution of network architectures: smaller convolutional kernel size, smaller convolutional stride, more convolutional channel, deeper network structure. In this paper, we choose the VGGNet-19 as our main investigated structure due to its good performance in object recognition, which is composed of 16 convolutional layers and 3 fully connected layers. The detailed description about VGGNet-19 is out of the scope of this paper and can be found in \cite{SimonyanZ14a}.

\textbf{Data augmentations.} By data augmentation, we mean perturbing an image by transformations that leave the underlying class unchanged. Typical transformations include corner cropping, scale jittering, and horizontal flipping. Specifically, during the training phase of OS-CNNs, we randomly crop image regions ($224 \times 224$) from 4 corners and 1 center of the whole image. Meanwhile these cropped regions undergo horizontal flipping randomly. Furthermore, we use three different scales to resize training images, where the smallest size $s$  of an image is set to $256, 384, 512$.

It should be noted that data augmentation is a method applicable to both training images and testing images.  During training phase, data augmentation will generate additional training examples and reduce the influence of over-fitting. For testing phase, data augmentation will help to improve the classification accuracy. The augmented samples can be either regarded as independent images or combined into a single representation by pooling or stacking operations. In the current implementation, during the test phase, we use sum pooling to aggregate these representations of augmented samples into a single representation.

\textbf{Learning policy.} Effective training methods are very crucial for learning CNN models. As the training dataset of cultural event recognition is relatively small compared with ImageNet \cite{DengDSLL009} and Places205 \cite{ZhouLXTO14}, we resort to pre-training OS-CNNs by using these public available models trained on ImageNet and Places205. Specifically, we pre-train object nets with public VGGNet-19 model \footnote{\url{http://www.robots.ox.ac.uk/~vgg/research/very_deep/}}, which achieved the top performance at ILSVRC2014.  For scene net, we use the model released by \cite{WangGH015} \footnote{\url{https://github.com/wanglimin/Places205-VGGNet}} to initialize the network weights, which has obtained the best performance on the Places205 dataset so far.

The network weights are learned using the mini-batch stochastic gradient descent with momentum (set to 0.9). At each iteration, a mini-batch of 256 images is constructed by random sampling. The dropout ratios for fully connected layers are set as $0.5$. As we pre-train network weights with ImageNet and Places205 models, we set a smaller learning rate for fine tuning OS-CNNs: learning rate starts with $10^{-3}$, decreases to $10^{-4}$ after 5K iterations, decreases to $10^{-5}$ after 10K iterations and the training process ends at 12K iterations. To speed up the training process,  we use a Multi-GPU extension version \cite{WangXW015} of Caffe toolbox \cite{JiaSDKLGGD14}, which is publicly available online \footnote{\url{https://github.com/yjxiong/caffe}}.

\section{Exploring OS-CNNs}
\label{sec:feature}

We have introduced the architectures and implementation details about OS-CNNs in the previous section. In this section, as shown in Figure \ref{fig:pipeline}, we will focus on describing the explorations of OS-CNN activations from different layers and try to improve the recognition  performance.

\subsection{Scenario 1: OS-CNN predictions}
The simplest way to utilize OS-CNNs for cultural event recognition is directly using the outputs (softmax layer) of CNN networks as final prediction results. Specifically, given an image $\mathbf{I}$, its recognition score is calculated as follows:
\begin{equation}
  s_{os}(\mathbf{I}) = \alpha_o s_o(\mathbf{I}) + \alpha_s s_s(\mathbf{I}),
\end{equation}
where $s_o(\mathbf{I})$ and $s_s(\mathbf{I})$ are the prediction scores of object nets and scene nets, $\alpha_o$ and $\alpha_s$ are the fusion weights of object nets and scene nets. In the current implementation, fusion weights are set to be equal for object nets and scene nets.

\subsection{Scenario 2: OS-CNN global representations with pre-training}
Another way to deploy OS-CNNs for cultural event recognition is to treat them as generic feature extractors and use them to extract the global representation of an image region. We usually extract the activations of \textbf{fully connected layers}, which are very compact and discriminative. In this case, we only use the pre-trained models without fine-tuning. Specifically, given an image region $I$, we extract this global representation based on OS-CNNs as follows:
\begin{equation}
 \phi^p_{os}(\mathbf{I}) = [\beta_o \phi^p_o(\mathbf{I}), \ \ \beta_s \phi^p_s(\mathbf{I})],
\end{equation}
where $\phi^p_o(\mathbf{I})$ and $\phi^p_s(\mathbf{I})$ are the CNN activations from pre-trained object nets and scene nets,  $\beta_o$ and $\beta_s$ are the fusion weights of object nets and scene nets. In current implementation, the fusion weights are set to be equal for object nets and scene nets.

\subsection{Scenario 3: OS-CNN global representations with pre-training and fine-tuning}

In previous scenario, OS-CNNs are only pre-trained on large scale dataset of object recognition and scene recognition, and directly applied to the smaller event recognition dataset. However, it was demonstrated that fine-tuning a pre-trained CNNs on the target data can improve the performance a lot \cite{GirshickDDM13}. We consider fine-tuning the OS-CNNs on the event recognition dataset and the resulted image representations become dataset-specific. After fine-tuning process, we obtain the following global representation with the fine-tuned OS-CNNs:
\begin{equation}
 \phi^f_{os}(\mathbf{I}) = [\beta_o \phi^f_o(\mathbf{I}), \ \ \beta_s \phi^f_s(\mathbf{I})],
\end{equation}
where $\phi^f_o(\mathbf{I})$ and $\phi^f_s(\mathbf{I})$ are the CNN activations from the fine-tuned object nets and scene nets,  $\beta_o$ and $\beta_s$ are the fusion weights of object nets and scene nets. In current implementation, the fusion weights are set to be equal for object nets and scene nets.

\subsection{Scenario 4: OS-CNN local representations + Fisher vector}
In previous two scenarios, we extract a global representation of an image region with OS-CNNs. Although this global representation is compact and discriminative, it may lack the ability of describing local patterns and detailed information. Inspired by the recent success on video-based action recognition with deep convolutional descriptors \cite{WangQT15a},  we investigate the effectiveness of \textbf{convolutional layer} activations. Convolutional layer features have been also demonstrated to be effective in image-based tasks, such as object recognition \cite{GaoWWL15}, scene recognition \cite{Dixit_2015_CVPR} and texture recognition \cite{CimpoiMKV15}. In this scenario, OS-CNNs are first pre-trained on large-scale ImageNet and Places205 datasets, and then fine-tuned on the event recognition dataset, just as in Scenario 3.

Specifically, given an image region $\mathbf{I}$, we first extract the convolutional feature maps of OS-CNNs (activations of convolutional layers) $C(\mathbf{I}) \in \mathbb{R}^{n \times n \times c}$, where $n$ is feature map size and $c$ is feature channel number. Each activation value in the convolutional feature map corresponds to a local receptive field in the original image, and therefore we call these activations of convolutional layers as OS-CNN local representations.

After extracting OS-CNN local representations, we utilize two normalization methods, namely \emph{channel normalization} and \emph{spatial normalization} proposed in \cite{WangQT15a}, to pre-process these convolutional feature maps into transformed convolutional feature maps  $\widetilde{C}(\mathbf{I}) \in \mathbb{R}^{n \times n \times c}$. More details regarding these two normalization methods are out scope of this paper and can be found in \cite {WangQT15a}. The normalized CNN activation $\widetilde{C}(\mathbf{I})(x,y,:) \in \mathbb{R}^{c}$ at each postion $(x,y)$ is called as the \emph{Transformed Deep-convolutional Descriptor} (TDD).  These two kinds of normalization methods have turned out to be effective for improving the performance of CNN local representations in \cite{WangQT15a}. Moreover, the combination of them can obtain higher performance. Therefore, we will use both normalization methods in our experimental explorations.

Finally, we employ Fisher vector \cite{SanchezPMV13} to encode these TDDs into a global representation due to its good performance in object recognition \cite{ChatfieldLVZ11} and action recognition \cite{SunN13a,WangWQ12}. In particular, according to our previous comprehensive study on encoding methods \cite{PengWWQ14}, we first use PCA to reduce the dimension of TDD to $64$. Then each TDD is soft-quantized with a Gaussian Mixture Model (GMM) with $K$ components ($K$ set to 256). The first and second order differences between each TDD $\mathbf{x} \in \mathbb{R}^{64}$ and its Gaussian center $\mu_k$ are aggregated in the block $\mathbf{u}_k$ and $\mathbf{v}_k$, respectively. The final Fisher vector representation is yielded by concatenating these blocks together:
\begin{equation}
 \phi_{fv}(\mathbf{I}) = [\mathbf{u}_1, \mathbf{v}_1, \cdots, \mathbf{u}_K, \mathbf{v}_K].
\end{equation}
For OS-CNNs,  the Fisher vector of local representation is defined as follows:
\begin{equation}
\phi_{os-fv}^f(\mathbf{I}) = [\beta_o \phi_{o-fv}^f(\mathbf{I}), \ \ \beta_s \phi_{s-fv}^f(\mathbf{I})],
\end{equation}
where $\phi_{o-fv}^f(\mathbf{I})$ is the Fisher vector representation from object nets, $\phi_{s-fv}^f(\mathbf{I})$ is the Fisher vector representation from scene nets, $\beta_o$ and $\beta_s$ are their fusion weights and set to be equal to each other in the current implementation.

\subsection{Linear classifiers}
All the representations $\phi(\mathbf{I})$ in previous three scenarios are used to construct a linear classifier $s(\mathbf{w}, \mathbf{I}) = \mathbf{w} \phi(\mathbf{I})$, where $\mathbf{w}$ is the weight of linear classifier. In our implementation, we choose LIBSVM \cite{ChangL11} as the classifier to learn the weight $\mathbf{w}$, where the parameter $C$, balancing regularizer and loss, is set as $1$. It is worth noting that all these representations are first normalized before fed into SVM for training. For OS-CNN global representations, we use $\ell_2$-normalization, and for OS-CNN local representations, we use intra normalization and power $\ell_2$-normalization.

\section{Experiments}
\label{sec:exp}

\begin{figure*}[t]
  \includegraphics[width=\textwidth]{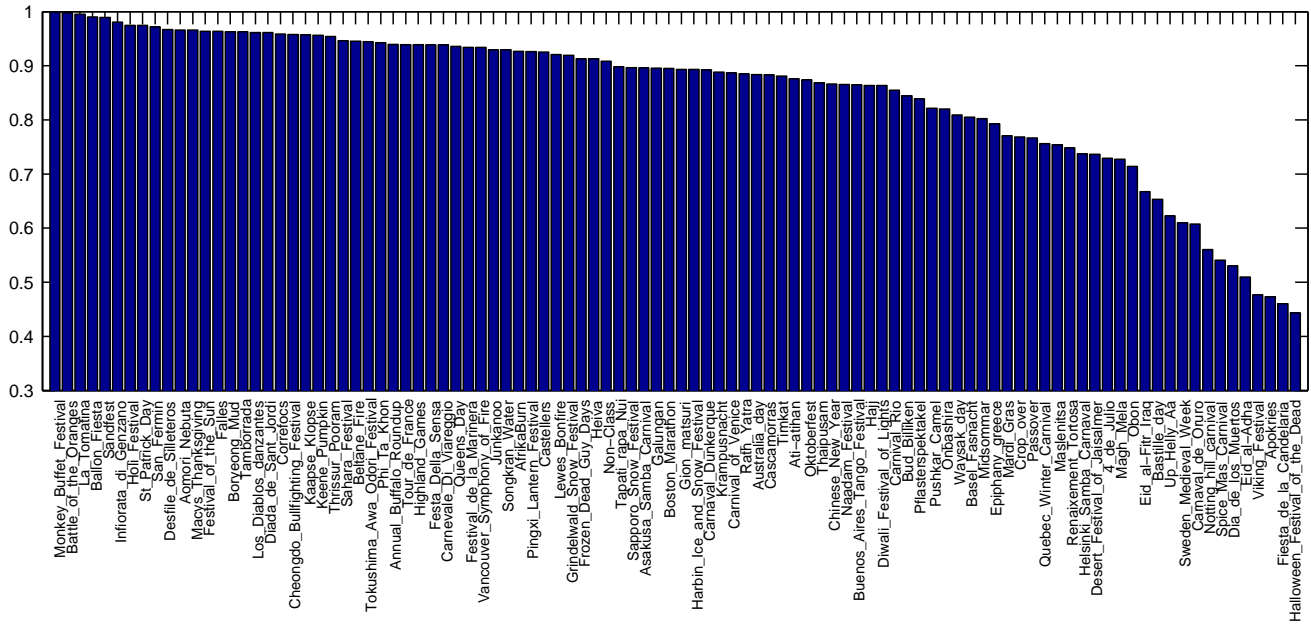}
  \caption{Per-class AP value of combining OS-CNN global and local representations on the validation data of ICCV ChaLearn LAP dataset.}
  \label{fig:ap}
\end{figure*}

\begin{table}[t]
\begin{tabular}{c|ccc}
\hline
& Object nets &  Scene nets & OS-CNNs \\ \hline
\textbf{Scenario 1} \\ \hline
softmax & 73.1\% & 71.2\% & 75.6\% \\ \hline
\textbf{Scenario 2} \\ \hline
fc7 & 67.2\% & 63.4\% & 69.1\% \\ \hline
\textbf{Scenario 3} \\ \hline
fc6 & 80.6\% & 76.8\% & 81.7\% \\ \hline
fc7 & 81.4\% & 78.1\% & 82.3\% \\ \hline
\textbf{Scenario 4} \\ \hline
conv5-1 &  77.6\% & 76.6\% & 78.9\% \\ \hline
conv5-2 & 78.6\% & 76.2\% & 79.6\% \\ \hline
conv5-3 & 79.4\% & 76.1\% & 80.2\% \\ \hline
conv5-4 & 78.4\% & 75.6\% & 79.7\% \\ \hline
\textbf{Fusion} \\ \hline
conv5-3+fc7 & 82.5\% & 79.3\% & 83.2\% \\ \hline
\end{tabular}
\vspace{4mm}
\caption{Event recognition performance of OS-CNN global and local representations on the validation data.}
\label{tbl:result}
\end{table}

In this section, we first describe the dataset of cultural event recognition at the  ICCV ChaLearn Looking at People (LAP) challenge 2015. Then we present and analyze the experimental results of our proposed different representations with OS-CNNs on the validation dataset of ChaLearn LAP dataset. Finally, we describe our solution for the ICCV ChaLearn LAP challenge 2015.

\begin{table}[t]
\centering
\begin{tabular}{|c|c|c|}
\hline
~~~Rank~~~ & ~~~~Team~~~~ & ~~~Score~~~\\
\hline
\hline
1 & VIPL-ICT-CAS & 85.4\% \\
2 & FV & 85.1\% \\
3 & \textbf{MMLAB (ours)} & 84.7\% \\
4 & NU\&C & 82.4\% \\
5 & CVL\_ETHZ& 79.8\% \\
6 & SSTK & 77.0\% \\
7 & MIPAL\_SUN & 76.3\% \\
8 & ESB & 75.8\% \\
9 & Sungbin Choi & 62.4\% \\
10 & UPC-STP & 58.8\% \\
\hline
\end{tabular}
\vspace{4mm}
\caption{Comparison the performance of our submission with those of other teams. Our team secures the third place in the ICCV ChaLearn LAP challenge 2015.}
\label{tbl:challenge}
\end{table}

\subsection{Datasets and evaluation protocol}
{\bf Datasets.} The ICCV  ChaLearn LAP challenge 2015 \cite{LAP15} contains a track of cultural event recognition and provides an event recognition dataset. This dataset contains images collected from two image search engines (Google Images and Bing Images). There are totally 100 event classes (99 event classes and 1 background class) from different countries and some images are shown in Figure \ref{fig:example}. From these samples, we see that cultural event recognition is really complicated, where garments, human poses, objects and scene context all constitute the possible cues to be exploited for event understanding. This dataset is divided into three parts: development data (14,332 images), validation data (5,704 images), and evaluation data (8,669 images). As we can not access the label of evaluation data, we mainly train our models on the development data and report the results on the validation data.

{\bf Evaluation protocol.} The principal quantitative measure is based on precision recall curve. They use the area under this curve as the computation of the average precision (AP), which is calculated by numerical integration. Finally, they average these per-class AP values across all event classes and employ the mean average precision (mAP) as the final ranking criteria. Hence, in our exploration experiments, we report our results evaluated as AP value for each class and mAP value for all classes.

\subsection{Results and analysis}
{\bf Settings.} In this exploration experiment, we use the VGGNet-19 as the OS-CNN network structure. We extract activations from two fully connected layers (\texttt{fc6}, \texttt{fc7}) as OS-CNN global representations, and activations from four convolutional layers (\texttt{conv5-1}, \texttt{conv5-2}, \texttt{conv5-3}, \texttt{conv5-4}) as OS-CNN local representations. It should be noted that we choose the activations after rectified Linear Units (ReLUs). We use $\ell_2$-normalization to further process OS-CNN global representations for better SVM training. For Fisher vector representation of OS-CNN local representation, we employ intra-normalization and power $\ell_2$-normalization, as suggested by \cite{PengWWQ14}.

{\bf Analysis.} We first report the numerical results in Table \ref{tbl:result}.  From these results, several conclusions can be drawn as follows:
\begin{itemize}
	\item We see that the object nets outperform scene nets on the task of cultural event recognition, which may imply that object cues play more important roles than scene cues for cultural event understanding.
	\item We observe that OS-CNNs are effective for event recognition as it extract both object and scene information from the image. They achieve superior performance to object nets and scene nets, no matter what scenario is adopted.
	\item We can notice that combining fine tuned features with linear SVM classifier (scenario 3) is able to obtain better performance than direct using the softmax output of CNNs (scenario 1). This result may be ascribed to the fact that CNNs are easily over-fitted to the training samples when the number of training images is relatively small.
	\item Comparing fine-tuned features (scenario 3) with pre-trained features (scenario 2), we may conclude that fine tuning on the target dataset is very useful for improving recognition performance, which agrees with the findings of 		\cite{GirshickDDM13}.
	\item Comparing the local representations (scenario 4) and global representations (scenario 3) of CNNs, we see that global representation achieve slightly higher recognition accuracy.
	\item We further combine the global representation (\texttt{fc7}) with  local representation (\texttt{conv5-3}) of CNNs and find that this combination is capable of boosting final recognition performance. This performance improvement indicates that different layers of CNNs capture different level abstraction of original image. These feature activations from different layers are complementary to each other.
\end{itemize}

\begin{figure*}[t]
  \includegraphics[width=\textwidth]{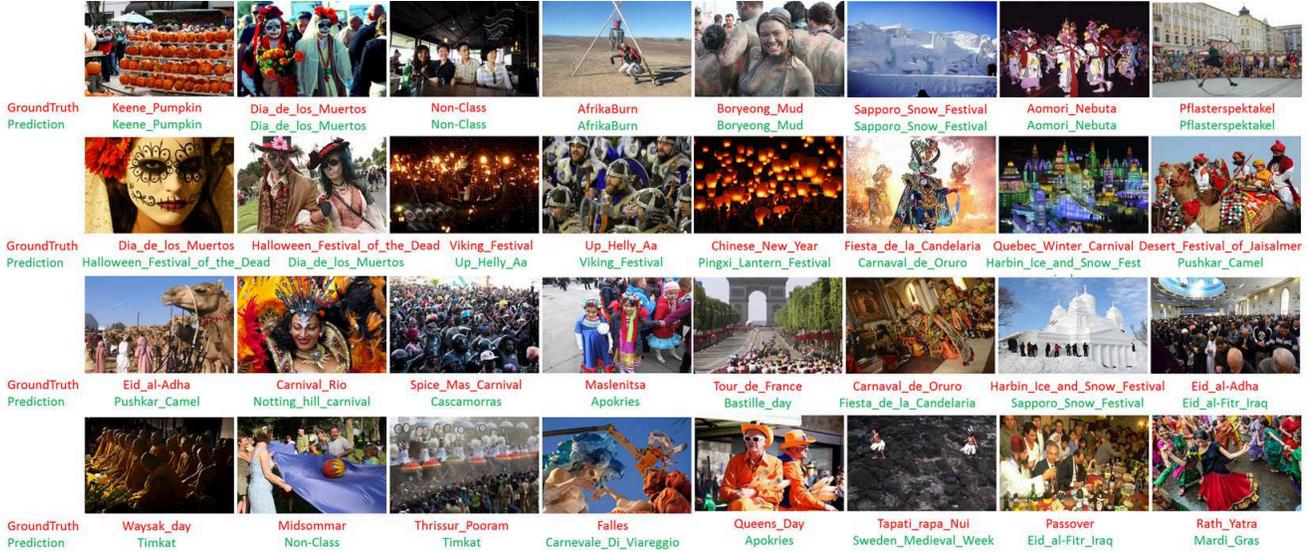}
 \caption{Examples of images that our method succeeds and fails in top-1 evaluation. We give 8 successfully predicted and 24 wrong predicted images in the row 1 and rows 2,3,4, respectively.}
  \label{fig:result_example}
  \vspace{-3mm}
\end{figure*}

We also plot the AP values for all event classes in Figure \ref{fig:ap}. From these AP values, we see that the events of \texttt{Monkey Buffet Festival} and \texttt{Battle of the Oranges} achieve the highest performance (100\%). This result may be ascribed to the fact that there are specific objects in these two event categories. At the same time, we notice that some event classes obtain very low AP values, such as \texttt{Halloween Festival of the Dead}, \texttt{Fiesta de la Candelaria}, \texttt{Apokries}, and \texttt{Viking Festival}. The AP values of these cultural event classes are below 50\%. In general, there are no specific objects and scene context in these difficult event classes, and besides these classes are easily confused with other classes from the perspective of visual appearance, as observed from Figure \ref{fig:result_example}.

We visualize several recognition examples in Figure \ref{fig:result_example}. In the row 1, we give eight examples that are successfully predicted by our method, from classes like \texttt{Keene Pummpking}, \texttt{Boryeong Mud}, \texttt{AfrikaBurn} and so on. Meanwhile, we also provide some failure cases with high confidence from our method in the rows 2,3,4. From these wrong predicted examples, we see that these failure cases are rather reasonable and there exists great confusion between some cultural event classes. For example, the event classes of \texttt{Dia de los Muertos} and \texttt{Halloween Festival of the Dead} share similar human make-up and garments. The event classes of \texttt{Up Helly Aa} and \texttt{Viking Festtival} share similar human dresses and containing objects. The event classes of \texttt{Harbin Icen and Snow Festival} and \texttt{Sapporo Snow Festival} share similar scene context and color appearance. The event classes of \texttt{Chinese New Year} and \texttt{Pingxi Lattern Festival} share similar containing objects. In summary, these examples in Figure \ref{fig:result_example} indicate that the concept of event is really complicated and there only exist slight difference between some event classes.

\subsection{Challenge results}
For final evaluation, we merge the development data (14,332 images) and validation data (5,704 images) into a single training dataset (20,036 images) and re-train our OS-CNN models on this new dataset. Our final submission results to the ICCV ChaLearn LAP challenge are based on our re-trained model.

According to the above experimental explorations, we conclude that the OS-CNN global and local representations are complementary to each other. Thus, we choose to combine activations from \texttt{fc7} and \texttt{conv5-3} layers, to keep a balance between performance and efficiency. Meanwhile, our previous study demonstrated that GoogLeNet is complementary to VGGNet \cite{WangWDQ15}. Hence, we also extract a global representation by using the OS-CNNs of GoogLeNet in our challenge solution. In summary, our challenge solution is composed of three representations: (i) OS-CNN VGGNet-19 local representations, (ii) OS-CNN VGGNet-19 global representations, and (iii) OS-CNN GoogLeNet global representations.

The challenge results are summarized in Table \ref{tbl:challenge}. We see that our method is among the top performers and our mAP is very close to the best performance of this challenge (84.7\% vs. 85.4\%). Regarding computational cost, our implementation is based on CUDA 7.0 and Matlab 2013a, and it takes about 1s to process one image in our workstation equipped with 8 cores CPU, 48G RAM, and Tesla K40 GPU.

\section{Conclusions}
\label{sec:con}

In this paper, we have comprehensively studied different aspects of OS-CNNs for better cultural event recognition. Specifically, we investigate the effectiveness of CNN activations from different layers by designing four types scenarios of adapting OS-CNNs to the task of cultural event recognition. From our empirical study, we demonstrate that the CNN activations from convolutional layers and fully connected layers are complementary to each other, and the combination of them is able to boost recognition performance. Finally, we come up with a solution by using OS-CNNs at the ICCV ChaLearn LAP challenge and secure the third place. In the future, we may consider how to incorporate more visual cues such as human poses, garments, object and scene relationship in a systematic manner for event recognition in still images.

\section*{Acknowledgement}
 This work is supported by a donation of two Tesla K40 GPUs from NVIDIA Corporation. Meanwhile this work is partially supported by National Natural Science Foundation of China (91320101, 61472410), Shenzhen Basic Research Program (JCYJ20120903092050890, JCYJ20120617114614438, JCYJ20130402113127496), 100 Talents Program of CAS, and Guangdong Innovative Research Team Program (No.201001D0104648280).

\clearpage
{
\bibliographystyle{ieee}
\bibliography{lap}
}

\end{document}